\documentclass[sigconf]{acmart}

\usepackage{amsmath}
\AtBeginDocument{%
  \providecommand\BibTeX{{%
    \normalfont B\kern-0.5em{\scshape i\kern-0.25em b}\kern-0.8em\TeX}}}

\setcopyright{acmcopyright}
\copyrightyear{2018}
\acmYear{2018}
\acmDOI{10.1145/1122445.1122456}

\acmConference[Gecco '22]{Gecco '22: The Genetic and Evolutionary Computation Conference}{July 09--13, 2022}{Boston}
\acmBooktitle{Gecco '22: The Genetic and Evolutionary Computation Conference,
  July 09--13, 2022, Boston}
\acmPrice{15.00}
\acmISBN{978-1-4503-XXXX-X/18/06}

\usepackage{array}
\usepackage{multirow}

\renewcommand{\bar}{\overline}

\begin{document}

\title{Benchmarking Quality-Diversity Algorithms\\on Neuroevolution for Reinforcement Learning}

\author{Manon Flageat} 
\authornote{Both authors contributed equally to this research.}
\affiliation{%
  \institution{Imperial College London}
  \streetaddress{Exhibition Road}
  \city{London}
  \country{United Kingdom}
  \postcode{SW7 2AZ}
}
\email{manon.flageat18@imperial.ac.uk}

\author{Bryan Lim}
\authornotemark[1]
\affiliation{%
  \institution{Imperial College London}
  \streetaddress{Exhibition Road}
  \city{London}
  \country{United Kingdom}
  \postcode{SW7 2AZ}
}
\email{bryan.lim16@imperial.ac.uk}

\author{Luca Grillotti}
\affiliation{%
  \institution{Imperial College London}
  \streetaddress{Exhibition Road}
  \city{London}
  \country{United Kingdom}
  \postcode{SW7 2AZ}
}
\email{luca.grillotti16@imperial.ac.uk}

\author{Maxime Allard}
\affiliation{%
  \institution{Imperial College London}
  \streetaddress{Exhibition Road}
  \city{London}
  \country{United Kingdom}
  \postcode{SW7 2AZ}
}
\email{m.allard20@imperial.ac.uk}

\author{Sim\'on C. Smith}
\affiliation{%
  \institution{Imperial College London}
  \streetaddress{Exhibition Road}
  \city{London}
  \country{United Kingdom}
  \postcode{SW7 2AZ}
}
\email{s.smith-bize@imperial.ac.uk}

\author{Antoine Cully}
\affiliation{%
  \institution{Imperial College London}
  \streetaddress{Exhibition Road}
  \city{London}
  \country{United Kingdom}
  \postcode{SW7 2AZ}
}
\email{a.cully@imperial.ac.uk}

\renewcommand{\shortauthors}{AIRL lab}

\begin{abstract}
  We present a Quality-Diversity benchmark suite for Deep Neuroevolution in Reinforcement Learning domains for robot control. The suite includes the definition of tasks, environments, behavioral descriptors, and fitness.
We specify different benchmarks based on the complexity of both the task and the agent controlled by a deep neural network.
The benchmark uses standard Quality-Diversity metrics, including coverage, QD-score, maximum fitness, and an archive profile metric to quantify the relation between coverage and fitness.
We also present how to quantify the robustness of the solutions with respect to environmental stochasticity by introducing corrected versions of the same metrics. We believe that our benchmark is a valuable tool for the community to compare and improve their findings. The source code is available online 
\footnote{\url{https://github.com/adaptive-intelligent-robotics/QDax}}.
\end{abstract}

\ccsdesc[500]{Computing methodologies~Evolutionary robotics}

\keywords{Quality-Diversity, Evolutionary algorithms}



\begin{teaserfigure}
\begin{center}
\includegraphics[width = \hsize]{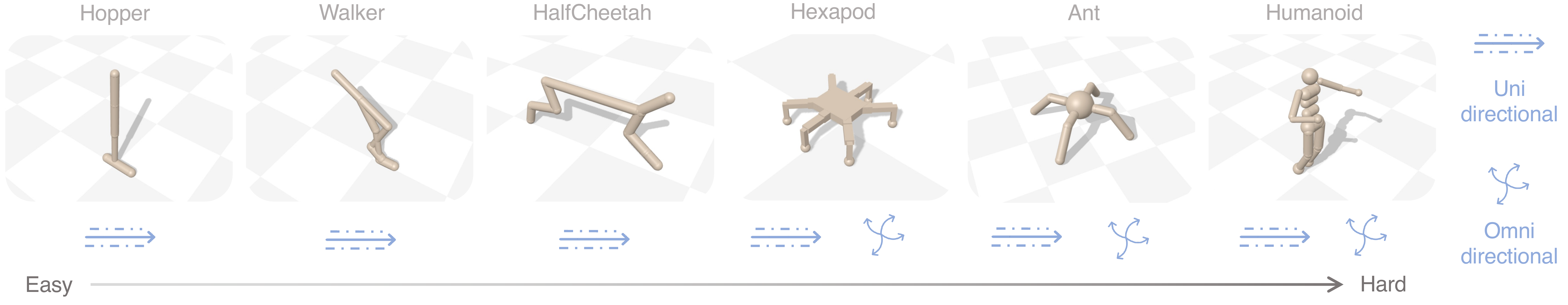}
\caption{
The 6 environments considered to define the uni-directional and omni-directional tasks. 
We provide an implementation of these environments and of the corresponding tasks in the QDax \cite{lim2022accelerated} library in the Brax simulator \cite{freeman2021brax}.
}
\label{fig:tasks}
\end{center}
\end{teaserfigure}
\maketitle

\newcommand{\archive}{\mathcal{A}}
\newcommand{\cardArchive}{{N_{\archive}}}
\newcommand{\cardinalFitness}{{N_{f}}}
\newcommand{\numberFit}[1]{n\left( f_{{#1}} \right)}
\newcommand{\numberSupFit}[1]{n_{\geq} \left( f_{{#1}} \right)}
\newcommand{\area}{a}
\newcommand{\fitnorm}{\widetilde{f}}

\section{Introduction}

\begin{table*}
  \begin{tabular}{ l | cccccc | ccc }
    \toprule
    \multirow{2}{*}{QD Task} & \multicolumn{6}{c}{Uni-directional} & \multicolumn{3}{c}{Omni-directional} \\
    \cmidrule(lr){2-7} \cmidrule(lr){8-10}
    &Hopper&Walker&Half-cheetah& Ant & Humanoid & Hexapod & Ant & Humanoid & Hexapod\\
    \midrule
    \addlinespace
    State Space & 14 & 20 & 26 & 87 & 299 & 169 & 87 & 299 & 169\\
    Action Space & 3 & 6 & 6 & 8 & 17 & 18 & 8 & 17 & 18 \\
    Parameters & $5{,}315$ & $5{,}894$ & $10{,}312$ & $24{,}465$ & $24{,}465$ & $16{,}210$ & $24{,}465$ & $24{,}465$ & $16{,}210$ \\
    BD Dimensions & 1 & 2 & 2 & 4 & 2 & 6 & 2 & 2 & 2\\
    Number of cells & $30^2$ & $30^2$ & $30^2$ & $5^4$ & $30^2$ & $5^6$ & $100^2$ & $100^2$ & $100^2$\\
    \bottomrule
\end{tabular}
  \caption{Summary of two tasks across six environments of different complexity, proposed to benchmark QD algorithms applied to neuroevolution for RL domains.
  In the last line, the notation $x^y$ gives the number of cells in the corresponding MAP-Elites archive, where $y$ corresponds to the BD dimensionality and $x$ to the number of sub-divisions of each BD dimension.}
  \label{tab:tasks}
\end{table*}

    
    

Quality-Diversity (QD) algorithms~\cite{chatzilygeroudis2021quality, pugh2016quality, cully2017quality} can be used to generate a collection of diverse and high-performing solutions. They have been used across many different domains such as robotics~\cite{cully2013behavioral, cully2015robots}, video game design~\cite{fontaine2020illuminating}, aerodynamic design~\cite{gaier2018data} and many others~\cite{fontaine2019mapping, ecoffet2021first}. 
Recent multiple works have considered applying QD algorithms to deep neuroevolution in Deep Reinforcement Learning (DRL) settings and showed competitive results \cite{ecoffet2021first, colas2020scaling, nilsson2021policy, pierrot2021diversity, tjanaka2022approximating}.
DRL methods conventionally rely on gradient-based optimization methods to search for optimal parameters of Deep Neural Networks (DNNs).
Conversely, the field of deep neuroevolution studies evolving DNNs with evolutionary methods where the genotypes are the parameters of the neural network~\cite{such2017deep, salimans2017evolution, risi2019deep}.

However, two key challenges arise when using QD algorithms for deep neuroevolution in RL domains; (i) the large number of parameters of DNNs making optimization in this high-dimensional search-space challenging, especially with conventional genetic mutations \cite{colas2020scaling, nilsson2021policy, gaier2020discovering}, and (ii) the stochasticity of these environments~\cite{flageat2020fast,colas2020scaling,nilsson2021policy}. 

DRL domains commonly consist of stochastic environments where initial states are sampled from distributions, and transitions between states may be non-deterministic. 
Furthermore, due to the use of closed-loop DNN policies, the stochasticity of the environments will result in deviations from states originally encountered and experienced by the policy. 
These deviations cause the policy to take different actions which result in varying behavioural descriptors and fitnesses for a single policy, making this another key challenge for QD algorithms in this context \cite{flageat2020fast,justesen2019map}.

The lack of standardised environments and tasks in QD for deep neuroevolution in RL domains makes quantifying scientific progress of the algorithms difficult.
While not perfect, benchmarks have been critical to progress in machine learning.
The lack of benchmarks has been most evident in fields such as computer vision~\cite{deng2009imagenet, lin2014microsoft, barbu2019objectnet} and speech recognition~\cite{panayotov2015librispeech}.
Deep RL also has benefited from benchmarking environments and platforms such as OpenAI Gym~\cite{brockman2016openai} and Deepmind Control Suite~\cite{tassa2018deepmind}.
In this paper, we propose a benchmark composed of two tasks for neuroevolution in RL domains across six different environments of varying complexity. 
We also formalize existing evaluation procedures and metrics from the literature to account for the two key challenges when evaluating algorithms in these domains: high-dimensional search-space and stochasticity.

%
%
\section{Tasks} \label{sec:tasks}

The benchmark we propose contains two tasks across six robot control environments, which are already included in the Brax physics simulator~\cite{freeman2021brax} except for the Hexapod environment. 
We propose several levels of difficulty of tasks based on the different morphologies of the robots (see Fig. \ref{fig:tasks}).
2-dimensional planar robots like the Hopper, Walker and Half-Cheetah have fewer degrees of freedom and a smaller state space making them easier to control.
More complex 3-dimensional robots like the Ant and the Humanoid have a larger number of degrees of freedom and larger state-space. In the case of the Humanoid, it is also unstable in its standing position.
The unstable dynamics require some degree of balance control on top of the locomotion control, thus making the task more difficult.
We also use a simulated hexapod that corresponds to a real robot, similar to an open-sourced 18 DoF Hexapod \footnote{\url{https://github.com/adaptive-intelligent-robotics/Hexapod_Design_V2}} and previously used in the QD literature~\cite{cully2015robots,chatzilygeroudis2018reset}. Environments like this provide further opportunities for policies to be transferred into real-world systems, which could uncover further challenges and limitations of QD algorithms.

We propose two types of tasks: a uni-directional and an omni-directional locomotion task. 
The uni-directional task was first introduced in ~\citet{cully2015robots} while the omni-directional task was introduced in the BR-Evolution algorithm~\cite{cully2013behavioral}. 
These tasks have been extended to different continuous control systems that were part of the OpenAI Gym~\cite{brockman2016openai} RL benchmark in QDGym~\cite{colas2020scaling,nilsson2021policy}.
Both tasks are important for the benchmark since previous works have demonstrated that algorithms behave different in their abilities to explore the behavioural space and finding high-performing solutions~\cite{pierrot2021diversity,nilsson2021policy}.
A summary of all the proposed benchmarks tasks and environments can be found in Table~\ref{tab:tasks}. 
For all experiments we use QDax~\cite{lim2022accelerated}, an accelerated QD library\footnote{\url{https://github.com/adaptive-intelligent-robotics/QDax}} that utilizes the Brax~\cite{freeman2021brax} simulator for the environment implementations.

\subsection{Uni-directional tasks}

The goal of uni-directional locomotion tasks is to find diverse gaits for walking forward as fast as possible. 
The behavioral descriptor for this task is the average time each foot is in contact with the ground. This descriptor promotes diversity in locomotion gaits that are useful for adaptation to mechanical damage \cite{cully2015robots}.
The contact of each foot $C_i$ with the ground is a good estimate of how much each leg $l_i$ contributes to the locomotion.
The fitness function $f_{uni}$ is a sum of the $x$-displacement ($r_{forward}$), a survival bonus ($r_{survive}$) and the negative sum of the torques ($-r_{torque}$) at each time-step.
\begin{align}
    bd_{uni} &= \frac{1}{T} \sum_{t}^{T}{
    \begin{pmatrix}
    C_1(t) \\
    \vdots \\
    C_I(t) 
    \end{pmatrix}\textrm{, where $I$ is the number of feet.}
    } \\
    f_{uni} &= \sum_{t=0}^{T}{r_{forward} + r_{survive} + (-r_{torque})}
\end{align}


\subsection{Omni-directional tasks}

In the omni-directional locomotion task, the goal is to move efficiently in every direction.
 This task requires policies that result in diverse final positions of the centre of mass (CoM) of the robot. 
The behavioral descriptor is then defined as the final cartesian coordinate of the CoM at the end of the episode:
\begin{align}
    bd_{omni} &= 
    \begin{pmatrix}
        x_{T} \\
        y_{T}
    \end{pmatrix} \\
    f_{omni} &= \sum_{t=0}^{T}{r_{survive} + (-r_{torque})}
\end{align}
The fitness function $f_{omni}$ is identical to $f_{uni}$ without the displacement term $r_{forward}$.
\section{Metrics} \label{sec:metrics}

We consider metrics to evaluate (1) the efficiency of QD algorithms, (2) the properties of the returned archives, and (3) the robustness of the archives w.r.t. stochastic environments.

\subsection{Evaluating the Efficiency of QD Algorithms} \label{sec:metrics_usual}

In this section, we consider a QD algorithm that uses a MAP-Elites grid or an unstructured archive, as most previously introduced QD approaches \cite{cully2017quality, chatzilygeroudis2021quality}.
Such algorithms are usually compared based on three core archive properties:

\begin{itemize}
    \item \textit{Coverage}: the number $\cardArchive$ of individuals in an archive $\archive$ (or the number of filled cells when the container is a grid). It estimates the diversity present in $\archive$: $$\textrm{Coverage}(\archive) = \cardArchive$$
    
    
    \item \textit{QD Score}~\cite{pugh2016quality}:
    %
    %
    the sum of the normalized finesses of all individuals in an archive $\archive$. It evaluates both the performance and diversity of $\archive$:
    $$\textrm{QDScore}(\archive) = \sum_{i=1}^{\cardArchive}\frac{f_i - \textrm{min}_j(f_j)}{\textrm{max}_j(f_j) - \textrm{min}_j(f_j)}$$
    %
    
    \item \textit{Max Fitness}: the highest achieved fitness among all individuals present in an archive $\archive$: $$\textrm{MaxFitness}(\archive) = \textrm{max}_{1 \leq j \leq \cardArchive}(f_j)$$
\end{itemize}

The above metrics are often considered w.r.t. the number of evaluations. 
However, this comparison may not highlight the difference in time-complexity between the algorithms.
Recent works on accelerating QD algorithms \cite{lim2022accelerated} have raised awareness of the importance of real-time-efficiency in QD optimization.
Thus, we emphasize the importance of reporting each metric w.r.t time or the full execution time of each algorithm.

\subsection{Studying Archive Properties} \label{sec:metrics_profile}

%
%
While the metrics defined above can be used to analyze the evolution of the archive performance, they only give a partial view of the content of the archive. For instance, two algorithms might generate archives with similar QD Scores, while one will have a different Coverage or a higher Max Fitness. 
To simplify the comparison of these different metrics in a single figure, we consider a plot that we call the \textit{Archive Profile}.
Archive Profiles have been used in prior work~\cite{fontaine2021differentiable} to analyze QD archives. They represent on the $x$-axis a threshold fitness value $f_{\textrm{threshold}}$, and on the $y$-axis the coverage obtained when filtering out individuals whose score is lower than the threshold fitness:
\begin{align*}
\label{eq:number_sup_fit}
    \textrm{ArchiveProfile}(f_{\textrm{\textrm{threshold}}}) = \sum_{i=1}^{\cardArchive} \phi \left( f_i \geq f_{\textrm{threshold}} \right) \\ \text{ where } \forall P, \phi (P) = \begin{cases} 1 \text{ if $P$ is true} \\ 0 \text{ otherwise} \end{cases}
\end{align*}

\begin{figure}[t]
\begin{center}
\includegraphics[width = \hsize]{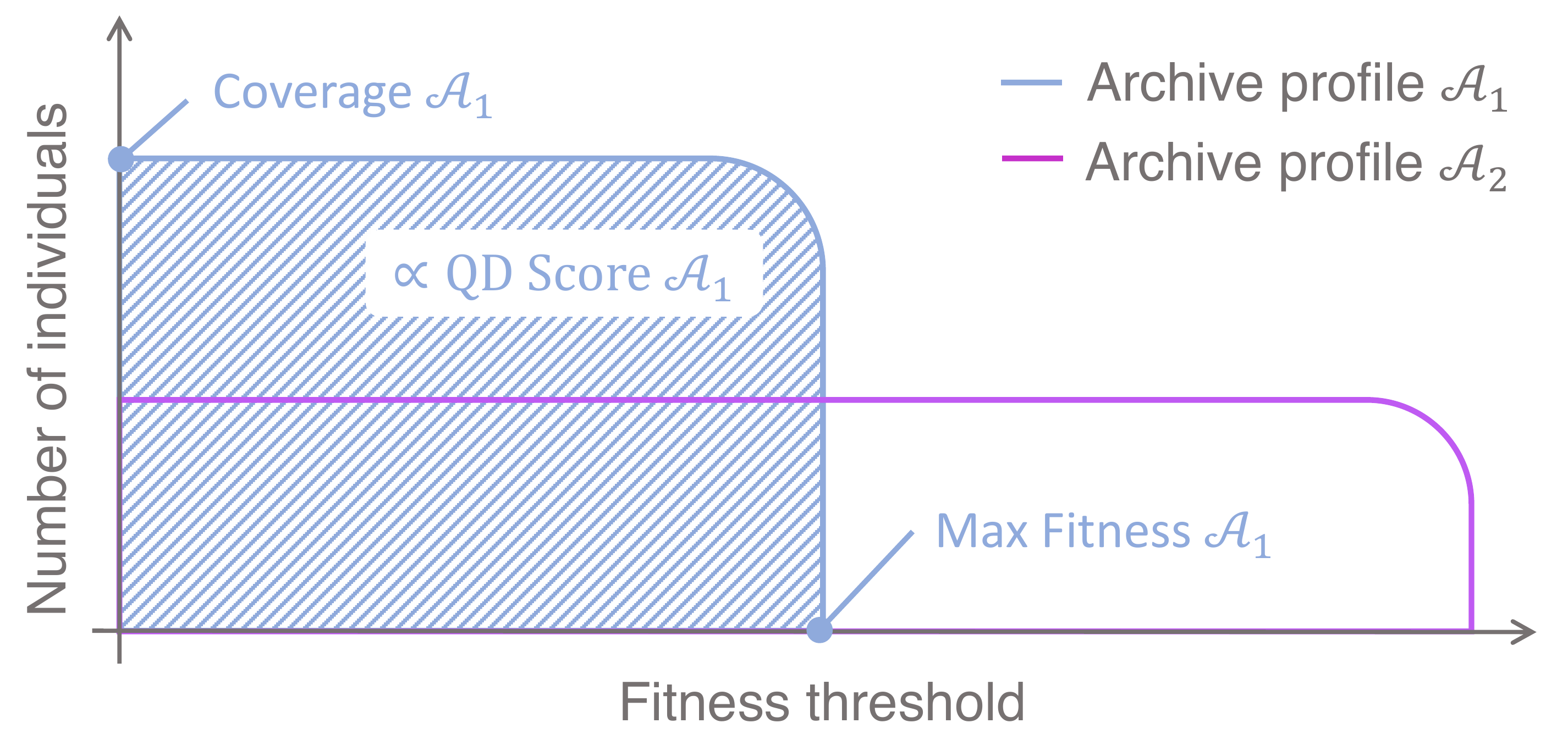}
\caption{
Comparison of the Archive Profile of two hypothetical archives $\archive_1$ and  $\archive_2$ with same QD Score (i.e. area under the curve) but different Archive Profile.
}
\label{fig:archive_profile}
\end{center}
\end{figure}

The function represented in an Archive Profile (see Fig.~\ref{fig:archive_profile}) expresses several properties and relationships with the three metrics from Section \ref{sec:metrics_usual}.
First of all, this function is monotonically decreasing and there is a fitness threshold $f^*$ such that: for all $f \geq f^*,\: \textrm{ArchiveProfile}(f)=0$.
That fitness threshold $f^*$ is the Max Fitness of the archive; and the highest value of $\textrm{ArchiveProfile}$ corresponds to the Coverage of the archive. 
Moreover, the area under the archive profile curve is proportional to the QD Score (see Appendix for proof).
That means the archive profile captures more information than the QD Score: archives having different QD Scores will exhibit different archive profiles, but two different archive profiles can have same QD Score (illustrated in Figure \ref{fig:archive_profile}).
%

\begin{figure*}[t]
\begin{center}
\includegraphics[width = \hsize]{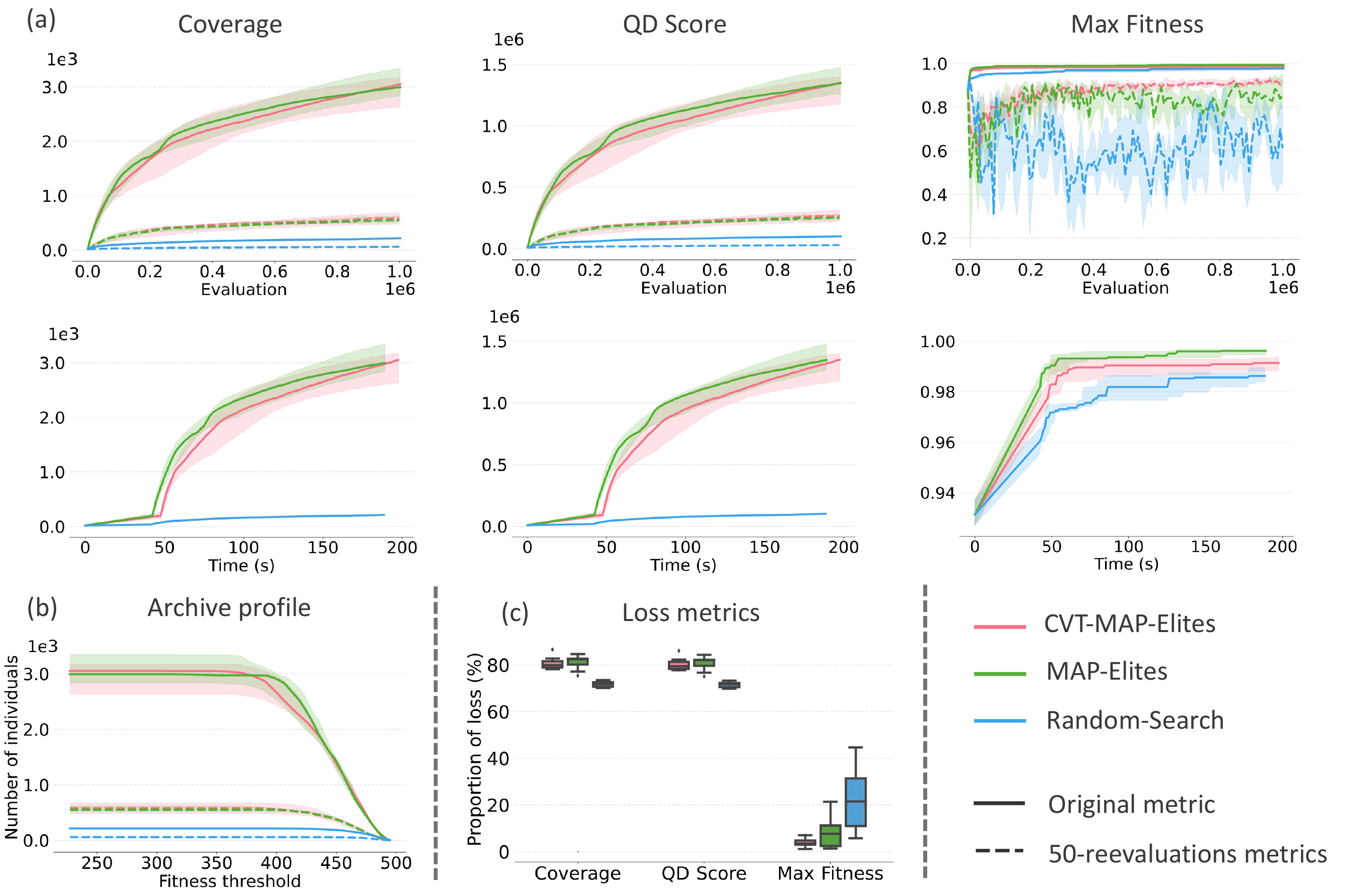}
\caption{
Comparison of MAP-Elites \cite{mouret2015illuminating}, CVT-MAP-Elites \cite{vassiliades2017using} and Random Search on the Ant-omni-directional task introduced in this paper. This figure displays all the metrics proposed in this work: (a) Coverage and Corrected Coverage (left), QD Score and Corrected QD Score (middle) and Max Fitness and Corrected Max Fitness (right) against number of evaluations (top) and time (bottom), (b) Archive profile and (c) Losses induced by stochasticity. 
}
\label{fig:graphs}
\end{center}
\end{figure*}

\subsection{Studying the Robustness to Stochasticity} \label{sec:metrics_stochasticity}

Stochasticity is a key challenge when applying QD to deep neuroevolution. 
Each solution kept by a QD algorithm is evaluated once during the optimisation process to compute its fitness and BD, but this unique estimation is subject to stochasticity and might be inaccurate. 
Hence, the metrics defined in Section \ref{sec:metrics_usual} and \ref{sec:metrics_profile} also rely on this single estimation, subjecting them to the same inaccuracies.
Here, we propose an expected fitness $\bar{f_i}$ and expected BD $\bar{bd_i}$ for each solution kept by the algorithm as the average over a fixed number of reevaluations $N$:
\begin{equation*}
\bar{f_i} = \frac{1}{N} \sum_{\textrm{eval}=1}^N f_i \quad \text{ and } \quad \bar{bd_i} = \frac{1}{N} \sum_{\textrm{eval}=1}^N bd_i
\end{equation*}
Note that the algorithms do not have access to these expected values during the optimization process. These reevaluations are only used for the computation of metrics and do not count toward the evaluation budget of the algorithms.
Most previous works considered $N=50$ \cite{justesen2019map, flageat2020fast}.
We place the reevaluated solutions in a "Corrected archive" $\bar{\archive}$ based on these estimations and following the usual archive-addition rules. 
The metrics characterizing this Corrected archive $\bar{\archive}$ constitute better estimations of the actual quality and diversity of the final collection. 
We refer to them as \textit{Corrected Coverage}, \textit{Corrected QD Score}, \textit{Corrected Max Fitness} and \textit{Corrected Archive profile}.

To further quantify the impact of stochasticity on the performance of QD algorithms in stochastic contexts, we also propose to study the loss induced by the reevaluation process:

$$\textrm{Loss}_{\textrm{Coverage}}(\archive) = \frac{\textrm{Coverage}(\archive) - \textrm{Coverage}(\bar{\archive})}{\textrm{Coverage}(\archive)}$$
$$\textrm{Loss}_{\textrm{QDScore}}(\archive) = \frac{\textrm{QDScore}(\archive) - \textrm{QDScore}(\bar{\archive})}{\textrm{QDScore}(\archive)}$$
$$\textrm{Loss}_{\textrm{MaxFitness}}(\archive) = \frac{\textrm{MaxFitness}(\archive) - \textrm{MaxFitness}(\bar{\archive})}{\textrm{MaxFitness}(\archive)}$$

A high-loss value indicates that the algorithm produces archives of individuals whose performance is not robust to stochasticity. These quantities should thus be minimized.
We refer to these metrics as \textit{Coverage Loss}, \textit{QD Score Loss} and \textit{Max Fitness Loss}.

\section{Results and Discussion}

Figure~\ref{fig:graphs} shows an example plot of all the metrics discussed in Section~\ref{sec:metrics} for the Omni-directional Ant task defined in Section~\ref{sec:tasks}. 
We study the performance of two well-established QD algorithms: MAP-Elites \cite{mouret2015illuminating} and CVT-MAP-Elites \cite{vassiliades2017using}, and we also consider Random Search as an additional baseline for comparison purposes. 
Each experiment is replicated $10$ times and we use $50$ reevaluations to compute the Corrected metrics. All baselines and replications use the same machine to make time comparisons fair. 
The two QD baselines perform similarly according to the usual QD metrics (Section~\ref{sec:metrics_usual}), and both outperform Random-Search, which is expected. 
However, the time-comparison indicates that CVT-MAP-Elites require additional time to reach similar performance, which is expected as the CVT mechanism require more computation time to determine the cell of an individual based on its bd.  While the difference, in this case, stays minimal, this emphasizes the importance of considering real-time to compare QD algorithms. 
These two baselines also have similar Archive profiles, significantly better than the one of Random-search. 

The Corrected metrics (Section~\ref{sec:metrics_stochasticity}) show that MAP-Elites and CVT-MAP-Elites both produce archives of individuals whose performance is not robust to stochasticity. The values of the Loss metrics are high for these two algorithms, reaching $80\%$ loss in both Coverage and QD Score. Interestingly, the few solutions discovered by Random search are more robust according to the Corrected and Loss metrics.
An important problem in stochastic environments is what we refer to as "lucky" solutions: solutions that have been evaluated with a score higher than the average score they can expect to get in this environment. 
Random search has no mechanism to favor promising individuals, thus it does not favor individuals that have been lucky. On the contrary, MAP-Elites and CVT-MAP-Elites tend to maintain such individuals, which migh indirectly enforce lucky individuals over truly high-performing ones.
Finally, the Corrected Max Fitness plots show that this metric has higher variations than the other Corrected metrics. 
The Max Fitness is computed on one-solution only, while the QD Score and Coverage rely on all solutions in the archive, which explains this higher variance.
These results demonstrate that stochasticity remains a key challenge for QD algorithms applied to neuroevolution in RL domains.

These first results illustrate the challenges of our proposed benchmark. They also demonstrate the importance of the full range of metrics introduced to evaluate algorithms performance on these tasks.
Still, an interesting direction of extension for this benchmark would be to broaden its focus by adding other types of environments such as manipulation tasks.

\begin{acks}
This work was supported by the Engineering and Physical Sciences Research Council (EPSRC) grant EP/V006673/1 project REcoVER. 
\end{acks}

\appendix

\section{Relationship between QD Score and the Area under Archive Profile}

We consider an archive $\archive$ with a finite number of individuals $\cardArchive$.
We write $(f_i)_{1\leq i \leq \cardinalFitness}$, all the \textit{distinct} fitness scores of individuals present in $\archive$, sorted in ascending order:
\begin{equation*}
    f_{min} \leq f_1 < f_2 < \ldots < f_\cardinalFitness \leq f_{max} 
\end{equation*}
where $[f_{min}, f_{max}]$ is a fixed interval including all fitness scores obtained by all individuals from all variants.

For all $i$, we write $\numberFit{i}$ the number of individuals whose fitness is equal to $f_i$.
Finally, $\numberSupFit{i}$ denotes the number of grid cells, whose fitness is greater or equal to $f_i$.
Thus:
\begin{equation}
\label{eq:number_sup_fit}
    \numberSupFit{i} = \sum_{j=i}^{\cardinalFitness} \numberFit{j}
\end{equation}
In other words, $\numberSupFit{i}$ is the coverage obtained if we only consider individuals whose fitness is higher or equal to $f_i$.

The area $\area$ under the archive profile curve can then be written:
\begin{align*}
    \area 
    &= \cardArchive (f_{1} - f_{min}) + \sum_{i=1}^{\cardinalFitness-1} (f_{i+1} - f_{i}) \numberSupFit{i}
    \\ &= \cardArchive (f_{1} - f_{min}) + \sum_{i=1}^{\cardinalFitness-1} \sum_{j=i}^{\cardinalFitness} \numberFit{j} (f_{i+1} - f_{i}) \tag*{see Eq.~(\ref{eq:number_sup_fit})}
    \\ &= \cardArchive (f_{1} - f_{min}) + \sum_{j=1}^{\cardinalFitness} \sum_{i=1}^{j-1} \numberFit{j} (f_{i+1} - f_{i})
    \\ &= \cardArchive (f_{1} - f_{min}) + \sum_{j=1}^{\cardinalFitness} \numberFit{j} (f_{j} - f_{1})
    \\ &= -\cardArchive f_{min} + \left( \sum_{j=1}^{\cardinalFitness} \numberFit{j} f_{j} \right) \tag*{as $\sum_{j=1}^{\cardinalFitness} \numberFit{j} = \cardArchive$}
\end{align*}

We now consider the normalised fitness scores $\fitnorm_i$, commonly used to compute the QD Score \cite{pugh2016quality, cully2017quality}:
\begin{align*}
    \forall i, \quad \fitnorm_i &= \frac{f_i - f_{min}}{f_{max} - f_{min}}
\end{align*}

Then, we obtain the following identity:
\begin{align*}
    \forall i, \quad f_i &= f_{min} + \fitnorm_i (f_{max} - f_{min})
\end{align*}

Thus, the area the under the archive profile curve can be expressed as follows:
\begin{align*}
    a &= (f_{max} - f_{min}) \underbrace{\left( \sum_{j=1}^{\cardinalFitness} \numberFit{j} \fitnorm_{j} \right)}_{\text{QD Score}}
\end{align*}
which shows that the area $a$ is proportional to the QD Score.

\bibliographystyle{ACM-Reference-Format}
\bibliography{biblio}

\end{document}